\documentclass{article}
\usepackage{ijcai17}

\usepackage{times}
\usepackage{amsmath}
\usepackage{multirow}
\usepackage{color}

\makeatletter
\newcommand{\@emptybiblabel}[1]{}
\makeatother
\usepackage{bm} 

\usepackage[english]{babel}

\usepackage{graphicx} 
\graphicspath{{figures/}} 
\usepackage{xcolor} 

\usepackage{booktabs}

\DeclareMathOperator*{\argmax}{arg\,max}

\usepackage[small]{caption}

\title{Bilateral Multi-Perspective Matching for Natural Language Sentences}
\author{Zhiguo Wang, Wael Hamza, Radu Florian \\ IBM T.J. Watson Research Center\\ { \{zhigwang,whamza,raduf\}@us.ibm.com}}

\begin{document}

\maketitle

\begin{abstract}
Natural language sentence matching is a fundamental technology for a variety of tasks. 
Previous approaches either match sentences from a single direction or only apply single granular (word-by-word or sentence-by-sentence) matching. In this work, we propose a bilateral multi-perspective matching (BiMPM) model. Given two sentences $P$ and $Q$, our model first encodes them with a BiLSTM encoder. Next, we match the two encoded sentences in two directions $P$ \textit{against} $Q$ and $Q$ \textit{against} $P$. In each matching direction, each time step of one sentence is matched against all time-steps of the other sentence from multiple perspectives. Then, another BiLSTM layer is utilized to aggregate the matching results into a fixed-length matching vector. Finally, based on the matching vector, a decision is made through a fully connected layer. We evaluate our model on three tasks: paraphrase identification, natural language inference and answer sentence selection. Experimental results on standard benchmark datasets show that our model achieves the state-of-the-art performance on all tasks. 
\end{abstract}

\section{Introduction}
Natural language sentence matching (NLSM) is the task of comparing two sentences and identifying the relationship between them. It is a fundamental technology for a variety of tasks. 
For example, in a paraphrase identification task, NLSM is used to determine whether two sentences are paraphrase or not~\cite{yin2015abcnn}. 
For a natural language inference task, NLSM is utilized to judge whether a hypothesis sentence can be inferred from a premise sentence~\cite{bowman2015large}. 
For question answering and information retrieval tasks, NLSM is employed to assess the relevance between query-answer pairs and rank all the candidate answers~\cite{wang2016sentence}. 
For machine comprehension tasks, NLSM is used for matching a passage with a question and pointing out the correct answer span~\cite{wang2016multi}.

With the renaissance of neural network models~\cite{lecun2015deep,peng2015circle,peng2016recurrent}, two types of deep learning frameworks were proposed for NLSM. The \textbf{first} framework is based on the ``Siamese'' architecture~\cite{bromley1993signature}. In this framework, the same neural network encoder (e.g., a CNN or a RNN) is applied to two input sentences individually, so that both of the two sentences are encoded into sentence vectors in the same embedding space. Then, a matching decision is made solely based on the two sentence vectors ~\cite{bowman2015large,tan2015lstm}. The advantage of this framework is that sharing parameters makes the model smaller and easier to train, and the sentence vectors can be used for visualization, sentence clustering and many other purposes~\cite{wang2016semi}. However, a disadvantage is that there is no explicit interaction between the two sentences during the encoding procedure, which may lose some important information. To deal with this problem, a \textbf{second} framework ``matching-aggregation'' has been proposed~\cite{wang2016compare,wang2016sentence}. Under this framework, smaller units (such as words or contextual vectors) of the two sentences are firstly matched, and then the matching results are aggregated (by a CNN or a LSTM) into a vector to make the final decision. The new framework captures more interactive features between the two sentences, therefore it acquires significant improvements. However, the previous ``matching-aggregation'' approaches still have some limitations. First, some of the approaches only explored the word-by-word matching~\cite{rocktaschel2015reasoning}, but ignored other granular matchings (e.g., phrase-by-sentence); Second, the matching is only performed in a single direction (e.g., matching $P$ against $Q$)~\cite{wang2015learning}, but neglected the reverse direction (e.g., matching $Q$ against $P$).

In this paper, to tackle these limitations, we propose a bilateral multi-perspective matching (BiMPM) model for NLSM tasks. Our model essentially belongs to the ``matching-aggregation'' framework. Given two sentences $P$ and $Q$, our model first encodes them with a bidirectional Long Short-Term Memory Network (BiLSTM). Next, we match the two encoded sentences in two directions $P \rightarrow Q$ and $P \leftarrow Q$. In each matching direction, let's say $P \rightarrow Q$, each time step of $Q$ is matched against all time-steps of $P$ from multiple perspectives. Then, another BiLSTM layer is utilized to aggregate the matching results into a fixed-length matching vector. Finally, based on the matching vector, a decision is made through a fully connected layer. We evaluate our model on three NLSM tasks: paraphrase identification, natural language inference and answer sentence selection. Experimental results on standard benchmark datasets show that our model achieves the state-of-the-art performance on all tasks.

In following parts, we start with a brief definition of the NLSM task (Section 2), followed by the details of our model (Section 3). Then we evaluate our model on standard benchmark datasets (Section 4). We talk about related work in Section 5, and conclude this work in Section 6.

\section{Task Definition}
\label{sec:task-def}
Formally, we can represent each example of the NLSM task as a triple $(P, Q, y)$,
where $P=(p_1, ..., p_j, ..., p_M)$ is a sentence with a length $M$, $Q=(q_1, ..., q_i, ..., q_N)$ is the second sentence with a length $N$, 
$y \in \mathcal{Y}$ is the label representing the relationship between $P$ and $Q$, and $\mathcal{Y}$ is a set of task-specific labels.
The NLSM task can be represented as estimating a conditional probability $\Pr{(y|P,Q)}$ based on the training set, 
and predicting the relationship for testing examples by 
$y^* = \argmax_{y \in \mathcal{Y}} \Pr(y|P, Q)$.
Concretely, for a paraphrase identification task, $P$ and $Q$ are two sentences, $\mathcal{Y}=\{0,1\}$, where $y=1$ means that $P$ and $Q$ are paraphrase of each other, and $y=0$ otherwise. 
For a natural language inference task, $P$ is a premise sentence, $Q$ is a hypothesis sentence, and $\mathcal{Y}=\{entailment, contradiction, neutral\}$ where $entailment$ indicates $Q$ can be inferred from $P$, $contradiction$ indicates $Q$ cannot be true condition on $P$, and $neutral$ means $P$ and $Q$ are irrelevant to each other. 
In an answer sentence selection task, $P$ is a question, $Q$ is a candidate answer, and $\mathcal{Y}=\{0,1\}$ where $y=1$ means $Q$ is a correct answer for $P$, and $y=0$ otherwise.

\section{Method}
In this section, we first give a high-level overview of our model in Sub-section \ref{subsec:overview}, and then give more details about our novel multi-perspective matching operation in Sub-section \ref{sub:multi-func}.

\subsection{Model Overview}
\label{subsec:overview}
\begin{figure}[tbp]
\begin{center}
\includegraphics[width=0.5\textwidth]{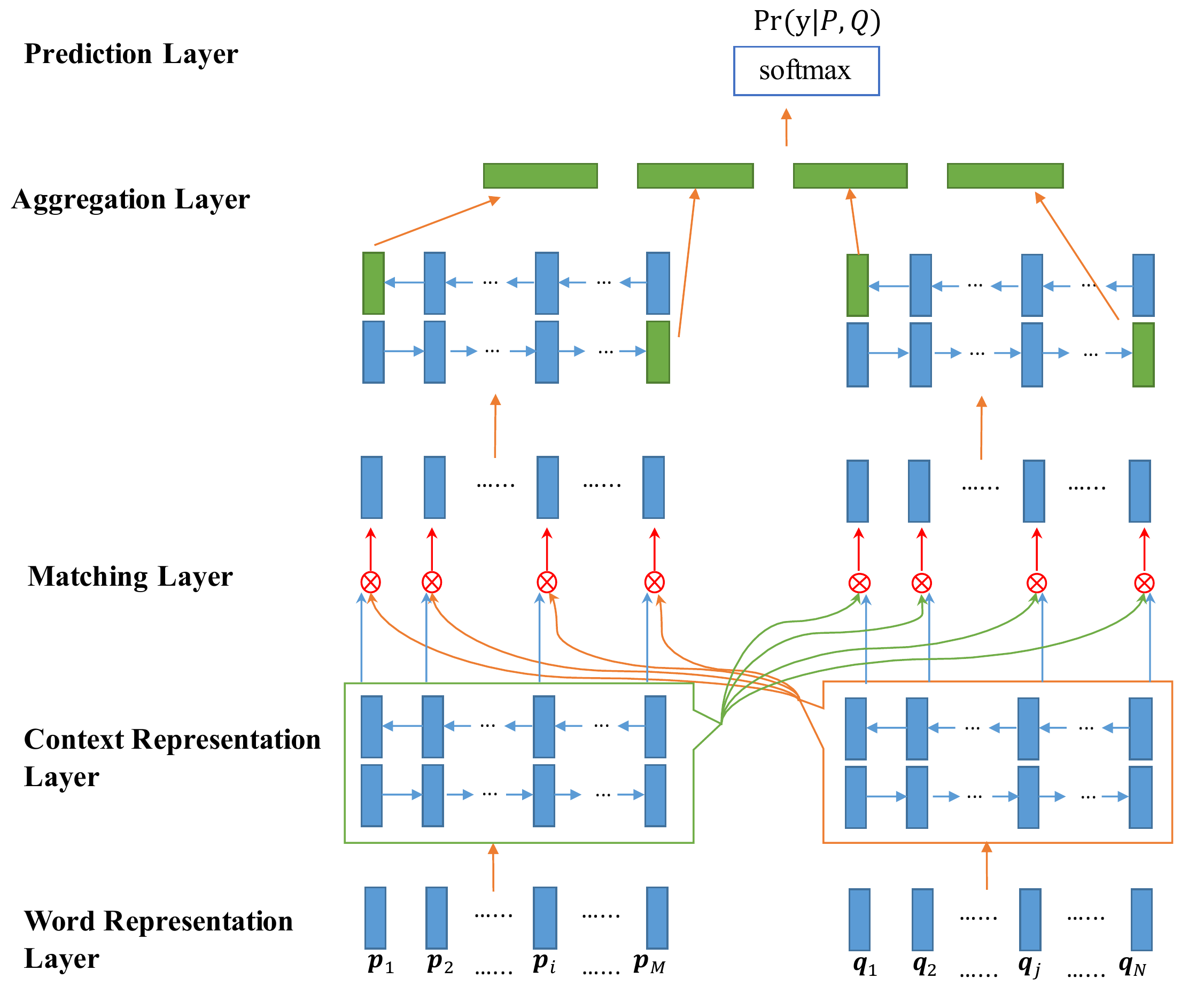}
\end{center}
\caption{Architecture for Bilateral Multi-Perspective Matching (BiMPM) Model, where  {\color{red} $\otimes$} is the multi-perspective matching operation described in sub-section \ref{sub:multi-func}.}
\label{fig:model-overview}
\end{figure}

We propose a bilateral multi-perspective matching (BiMPM) model to estimate the probability distribution $\Pr(y|P, Q)$. Our model belongs to the ``matching-aggregation'' framework~\cite{wang2016compare}. Contrarily to previous ``matching-aggregation'' approaches, our model matches $P$ and $Q$ in two directions ($P \rightarrow Q$ and $P \leftarrow Q$). In each individual direction, our model matches the two sentences from multiple perspectives. Figure \ref{fig:model-overview} shows the architecture of our model. Given a pair of sentences $P$ and $Q$, the BiMPM model estimates the probability distribution $\Pr(y|P, Q)$ through the following five layers.

\textbf{Word Representation Layer.} The goal of this layer is to represent each word in $P$ and $Q$ with a $d$-dimensional vector. 
We construct the $d$-dimensional vector with two components: a word embedding and a character-composed embedding. 
The word embedding is a fixed vector for each individual word, which is pre-trained with GloVe \cite{pennington2014glove} or word2vec \cite{mikolov2013distributed}. 
The character-composed embedding is calculated by feeding each character (represented as a character embedding) within a word into a Long Short-Term Memory Network (LSTM) \cite{hochreiter1997long}, where the character embeddings are randomly initialized and learned jointly with other network parameters from NLSM tasks. The output of this layer are two sequences of word vectors $P: [\textbf{\emph{p}}_1,...,\textbf{\emph{p}}_M]$ and $Q:[\textbf{\emph{q}}_1,...,\textbf{\emph{q}}_N]$.

\textbf{Context Representation Layer.} The purpose of this layer is to incorporate contextual information into the representation of each time step of $P$ and $Q$. We utilize a bi-directional LSTM (BiLSTM) to encode contextual embeddings for each time-step of $P$.
\begin{equation}
\begin{split}
\overrightarrow{\textbf{\emph{h}}}_i^p=\overrightarrow{\text{LSTM}}(\overrightarrow{\textbf{\emph{h}}}_{i-1}^p,\textbf{\emph{p}}_i) \hspace{10mm} i=1,...,M \\
\overleftarrow{\textbf{\emph{h}}}_i^p=\overleftarrow{\text{LSTM}}(\overleftarrow{\textbf{\emph{h}}}_{i+1}^p,\textbf{\emph{p}}_i) \hspace{10mm} i=M,...,1
\label{equ:biLSTM}
\end{split}
\end{equation}
Meanwhile, we apply the same BiLSTM to encode $Q$:
\begin{equation}
\begin{split}
\overrightarrow{\textbf{\emph{h}}}_j^q=\overrightarrow{\text{LSTM}}(\overrightarrow{\textbf{\emph{h}}}_{j-1}^q,\textbf{\emph{q}}_j) \hspace{10mm} j=1,...,N \\
\overleftarrow{\textbf{\emph{h}}}_j^q=\overleftarrow{\text{LSTM}}(\overleftarrow{\textbf{\emph{h}}}_{j+1}^q,\textbf{\emph{q}}_j) \hspace{10mm} j=N,...,1
\label{equ:biLSTM2}
\end{split}
\end{equation}

\textbf{Matching Layer.} 
This is the core layer within our model. 
The goal of this layer is to compare each contextual embedding (time-step) of one sentence against all contextual embeddings (time-steps) of the other sentence. As shown in Figure \ref{fig:model-overview}, we will match the two sentences $P$ and $Q$ in two directions: match each time-step of $P$ against all time-steps of $Q$, and match each time-step of $Q$ against all time-steps of $P$. To match one time-step of a sentence against all time-steps of the other sentence, we design a multi-perspective matching operation {\color{red} $\otimes$}. We will give more details about this operation in Sub-section \ref{sub:multi-func}. The output of this layer are two sequences of matching vectors (right above the operation {\color{red} $\otimes$} in Figure \ref{fig:model-overview}), where each matching vector corresponds to the matching result of one time-step against all time-steps of the other sentence.

\textbf{Aggregation Layer.} 
This layer is employed to aggregate the two sequences of matching vectors into a fixed-length matching vector. We utilize another BiLSTM model, and apply it to the two sequences of matching vectors individually. Then, we construct the fixed-length matching vector by concatenating (the four green) vectors from the last time-step of the BiLSTM models.

\textbf{Prediction Layer.} 
The purpose of this layer is to evaluate the probability distribution $\Pr(y|P, Q)$. To this end, we employ a two layer feed-forward neural network to consume the fixed-length matching vector, and apply the $softmax$ function in the output layer. The number of nodes in the output layer is set based on each specific task described in Section \ref{sec:task-def}.

\subsection{Multi-perspective Matching Operation}
\label{sub:multi-func}

We define the multi-perspective matching operation {\color{red} $\otimes$} in following two steps:

\textbf{First}, we define a multi-perspective cosine matching function $f_m$ to compare two vectors
\begin{equation}
\bm{m} = f_{m}(\bm{v}_1,\bm{v}_2;\bm{W})
\label{equ:MP_cosine}
\end{equation}
where $\bm{v}_1$ and $\bm{v}_2$ are two $d$-dimensional vectors, $\bm{W} \in \Re^{l \times d}$ is a trainable parameter with the shape $l \times d$, $l$ is the number of perspectives, and the returned value $\bm{m}$ is a $l$-dimensional vector $\bm{m}=[m_1,...,m_k,...,m_l]$. Each element $m_k \in \bm{m}$ is a matching value from the $k$-th perspective, and it is calculated by the cosine similarity between two weighted vectors
\begin{equation}
m_k=cosine(W_k \circ \bm{v}_1, W_k \circ \bm{v}_2)
\label{equ:weight_cosine}
\end{equation}
where $\circ$ is the element-wise multiplication, and $W_k$ is the $k$-th row of $\bm{W}$, which controls the $k$-th perspective and assigns different weights to different dimensions of the $d$-dimensional space.

\begin{figure}[tbp]
\begin{center}
\includegraphics[width=0.4\textwidth]{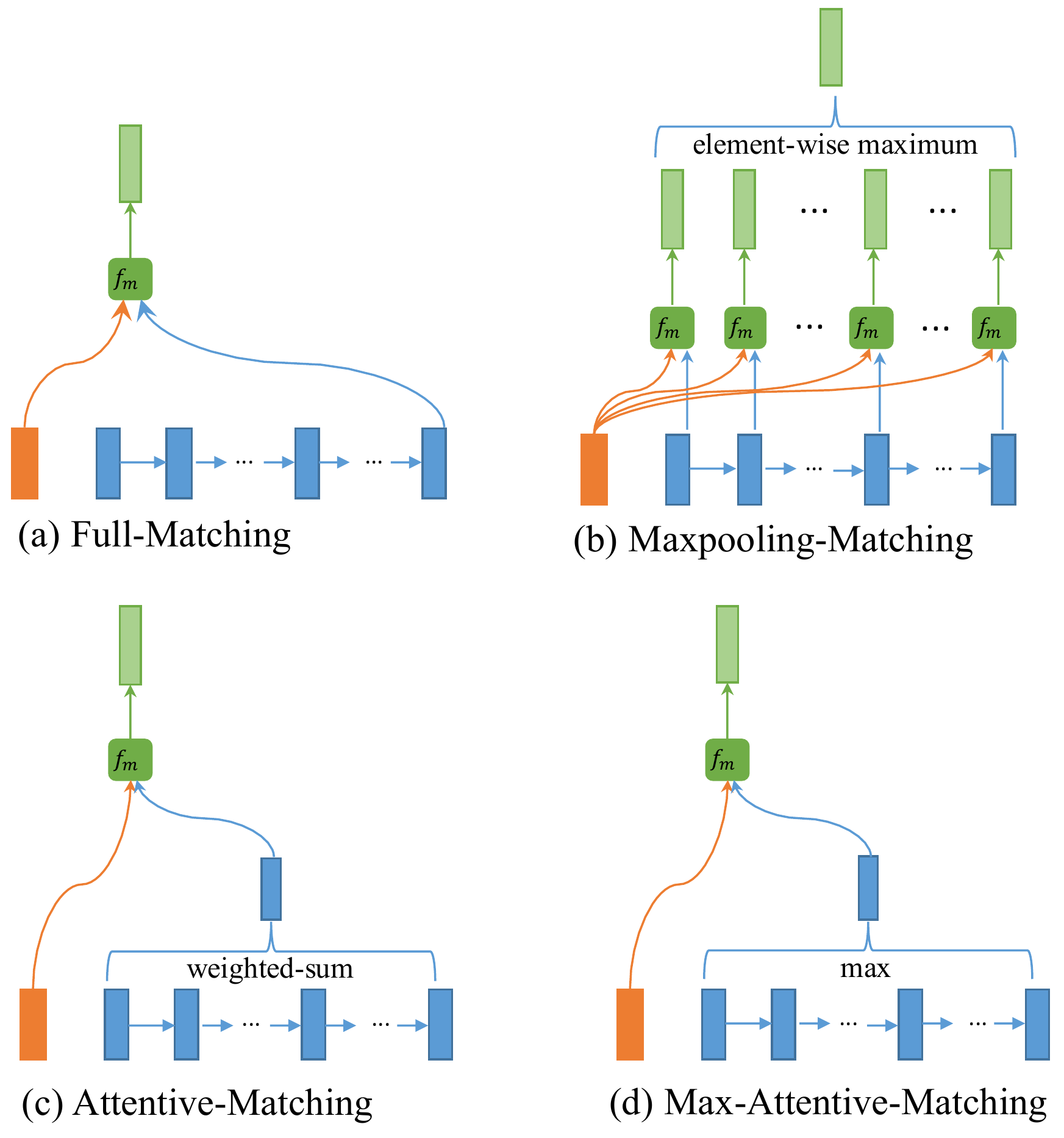}
\end{center}
\caption{Diagrams for different matching strategies, where $f_m$ is the multi-perspective cosine matching function in Eq.(\ref{equ:MP_cosine}), the input includes one time step of one sentence (left orange block) and all the time-steps of the other sentence (right blue blocks), and the output is a vector of matching values (top green block) calculated by Eq.(\ref{equ:MP_cosine}).}
\label{fig:match_strategies}
\end{figure}

\textbf{Second}, based on $f_{m}$, 
we define four matching strategies to compare each time-step of one sentence against all time-steps of the other sentence. To avoid repetition, we only define these matching strategies for one matching direction $P \rightarrow Q$. The readers can infer equations for the reverse direction easily.

(1) Full-Matching. Figure \ref{fig:match_strategies} (a) shows the diagram of this matching strategy. In this strategy, each forward (or backward) contextual embedding $\overrightarrow{\bm{h}}_i^p$ (or $\overleftarrow{\bm{h}}_i^p$) is compared with the last time step of the forward (or backward) representation of the other sentence $\overrightarrow{\bm{h}}_N^q$ (or $\overleftarrow{\bm{h}}_1^q$).
\begin{equation}
\begin{split}
\overrightarrow{\bm{m}}^{full}_i = f_{m}(\overrightarrow{\bm{h}}_i^p, \overrightarrow{\bm{h}}_N^q;\bm{W}^1) \\
\overleftarrow{\bm{m}}^{full}_i = f_{m}(\overleftarrow{\bm{h}}_i^p, \overleftarrow{\bm{h}}_1^q;\bm{W}^2) \\
\label{equ:FullM}
\end{split}
\end{equation}

(2) Maxpooling-Matching. Figure \ref{fig:match_strategies} (b) gives the diagram of this matching strategy. In this strategy, each forward (or backward) contextual embedding $\overrightarrow{\bm{h}}_i^p$ (or $\overleftarrow{\bm{h}}_i^p$) is compared with every forward (or backward) contextual embeddings of the other sentence $\overrightarrow{\bm{h}}_j^q$ (or $\overleftarrow{\bm{h}}_j^q$) for $j \in (1...N)$, and only the maximum value of each dimension is retained.

\begin{equation}
\begin{split}
\overrightarrow{\bm{m}}^{max}_i = \max_{j \in (1 ... N)} f_{m}(\overrightarrow{\bm{h}}_i^p, \overrightarrow{\bm{h}}_j^q;\bm{W}^3) \\
\overleftarrow{\bm{m}}^{max}_i = \max_{j \in (1 ... N)} f_{m}(\overleftarrow{\bm{h}}_i^p, \overleftarrow{\bm{h}}_j^q;\bm{W}^4) \\
\text{where $\max_{j \in (1 ... N)}$ is element-wise maximum.}
\label{equ:MaxM}
\end{split}
\end{equation}
 
(3) Attentive-Matching. Figure \ref{fig:match_strategies} (c) shows the diagram of this matching strategy. We first calculate the cosine similarities between each forward (or backward) contextual embedding $\overrightarrow{\bm{h}}_i^p$ (or $\overleftarrow{\bm{h}}_i^p$) and every forward (or backward) contextual embeddings of the other sentence $\overrightarrow{\bm{h}}_j^q$ (or $\overleftarrow{\bm{h}}_j^q$):

\begin{equation}
\begin{split}
\overrightarrow{\alpha}_{i,j}=cosine(\overrightarrow{\bm{h}}_i^p,\overrightarrow{\bm{h}}_j^q) \hspace{10mm} j=1,...,N \\
\overleftarrow{\alpha}_{i,j}=cosine(\overleftarrow{\bm{h}}_i^p,\overleftarrow{\bm{h}}_j^q) \hspace{10mm} j=1,...,N
\label{equ:attentions}
\end{split}
\end{equation}

Then, we take $\overrightarrow{\alpha}_{i,j}$ (or $\overleftarrow{\alpha}_{i,j}$) as the weight of $\overrightarrow{\bm{h}}_j^q$ (or $\overleftarrow{\bm{h}}_j^q$), and calculate an attentive vector for the entire sentence $Q$ by weighted summing all the contextual embeddings of $Q$:

\begin{equation}
\begin{split}
\overrightarrow{\bm{h}}_i^{mean}= \frac{\sum_{j=1}^N \overrightarrow{\alpha}_{i,j} \cdot \overrightarrow{\bm{h}}_j^q}{\sum_{j=1}^N \overrightarrow{\alpha}_{i,j}}\\
\overleftarrow{\bm{h}}_i^{mean}= \frac{\sum_{j=1}^N \overleftarrow{\alpha}_{i,j} \cdot \overleftarrow{\bm{h}}_j^q}{\sum_{j=1}^N \overleftarrow{\alpha}_{i,j}}\\
\label{equ:weighted_sum}
\end{split}
\end{equation}

Finally, we match each forward (or backward) contextual embedding of $\overrightarrow{\bm{h}}_i^p$ (or $\overleftarrow{\bm{h}}_i^p$) with its corresponding attentive vector:

\begin{equation}
\begin{split}
\overrightarrow{\bm{m}}^{att}_i = f_{m}(\overrightarrow{\bm{h}}_i^p,\overrightarrow{\bm{h}}_i^{mean};\bm{W}^5) \\
\overleftarrow{\bm{m}}^{att}_i = f_{m}(\overleftarrow{\bm{h}}_i^p, \overleftarrow{\bm{h}}_i^{mean};\bm{W}^6) \\
\label{equ:attentive_matching}
\end{split}
\end{equation}

(4) Max-Attentive-Matching. Figure \ref{fig:match_strategies} (d) shows the diagram of this matching strategy. This strategy is similar to the Attentive-Matching strategy. However, instead of taking the weighed sum of all the contextual embeddings as the attentive vector, we pick the contextual embedding with the highest cosine similarity as the attentive vector. Then, we match each contextual embedding of the sentence $P$ with its new attentive vector.

We apply all these four matching strategies to each time-step of the sentence $P$, and concatenate the generated eight vectors as the matching vector for each time-step of $P$. We also perform the same process for the reverse matching direction.

\section{Experiments}
\label{sec:exper}
In this section, we evaluate our model on three tasks: paraphrase identification, natural language inference and answer sentence selection. We will first introduce the general setting of our BiMPM models in Sub-section \ref{subsec:setting}. Then, we demonstrate the properties of our model through some ablation studies in Sub-section \ref{subsec:property}. Finally, we compare our model with state-of-the-art models on some standard benchmark datasets in Sub-section \ref{subsec:parahprase}, \ref{subsec:inference} and \ref{subsec:answer-selection}.

\subsection{Experiment Settings}
\label{subsec:setting}
We initialize word embeddings in the word representation layer with the 300-dimensional GloVe word vectors pre-trained from the 840B Common Crawl corpus \cite{pennington2014glove}. 
For the out-of-vocabulary (OOV) words, we initialize the word embeddings randomly.  
For the character-composed embeddings, we initialize each character as a 20-dimensional vector, and compose each word into a 50-dimensional vector with a LSTM layer. 
We set the hidden size as 100 for all BiLSTM layers.
We apply dropout to every layers in Figure \ref{fig:model-overview}, and set the dropout ratio as 0.1. 
To train the model, we minimize the cross entropy of the training set, and use the ADAM optimizer \cite{kingma2014adam} to update parameters. 
We set the learning rate as 0.001. 
During training, we do not update the pre-trained word embeddings. For all the experiments, we pick the model which works the best on the dev set, and then evaluate it on the test set.

\subsection{Model Properties}
\label{subsec:property}
To demonstrate the properties of our model, we choose the paraphrase identification task, and experiment on the ``Quora Question Pairs'' dataset \footnote{https://data.quora.com/First-Quora-Dataset-Release-Question-Pairs}. 
This dataset consists of over 400,000 question pairs, and each question pair is annotated with a binary value indicating whether the two questions are paraphrase of each other. 
We randomly select 5,000 paraphrases and 5,000 non-paraphrases as the dev set, and sample another 5,000 paraphrases and 5,000 non-paraphrases as the test set. We keep the remaining instances as the training set \footnote{We will release our source code and the dataset partition
at https://zhiguowang.github.io/
.}. 

\begin{figure}[tbp]
\begin{center}
\includegraphics[width=0.4\textwidth]{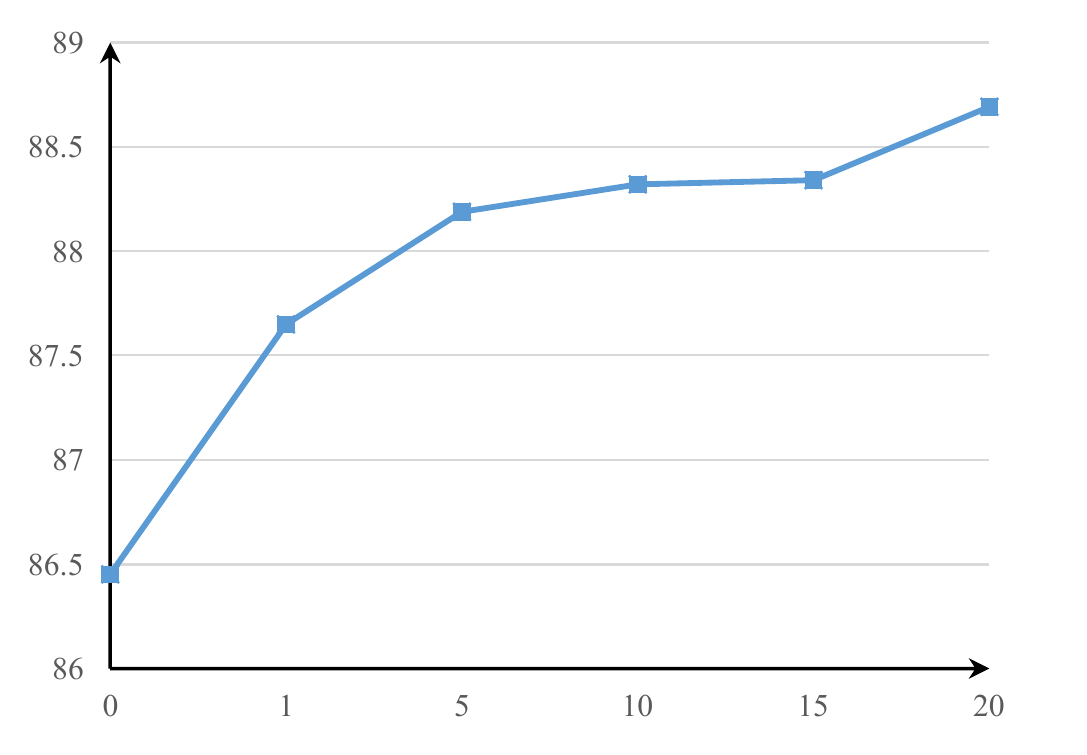}
\end{center}
\caption{Influence of the multi-perspective cosine matching function in Eq.(\ref{equ:MP_cosine}) .}
\label{fig:curve}
\end{figure}

First, we study the influence of our multi-perspective cosine matching function in Eq.(\ref{equ:MP_cosine}). 
We vary the number of perspectives $l$ among \{1, 5, 10, 15, 20\}\footnote{Due to practical limitations, we did not experiment with more perspectives.}, and keep the other options unchanged.
We also build a baseline model by replacing Eq.(\ref{equ:MP_cosine}) with the vanilla cosine similarity function.
Figure \ref{fig:curve} shows the performance curve on the dev set, where $l=0$ corresponds to the performance of our baseline model. 
We can see that, even if we only utilize one perspective ($l=1$), our model gets a significant improvement.
When increasing the number of perspectives, the performance improves significantly. Therefore, our multi-perspective cosine matching function is really effective for matching vectors. 

Second, to check the effectiveness of bilateral matching, we build two ablation models to matching sentences in only a single direction: 1) ``Only $P \rightarrow Q$'' which only matches $P$ against $Q$; 2) ``Only $P \leftarrow Q$'' which only matches $Q$ against $P$. Table \ref{tab:ablation} shows the performance on the dev set. Comparing the two ablation models with the ``Full Model'', we can observe that single direction matching hurts the performance for about 1 percent. Therefore, matching sentences in two directions is really necessary for acquiring better performance.

Third, we evaluate the effectiveness of different matching strategies. To this end, we construct four ablation models (w/o Full-Matching, w/o Maxpooling-Matching, w/o Attentive-Matching, w/o Max-Attentive-Matching) by eliminating a matching strategy at each time. Table \ref{tab:ablation} shows the performance on the dev set. We can see that eliminating any of the matching strategies would hurt the performance significantly.

\begin{table}[tbp]
\centering
\begin{tabular}{lc}
\toprule
Models                   & Accuracy\\
\midrule
Only $P \rightarrow Q$ & 87.74 \\
Only $P \leftarrow Q$ & 87.47 \\
\midrule
w/o Full-Matching       & 87.86 \\
w/o Maxpooling-Matching & 87.64 \\
w/o Attentive-Matching  & 87.87 \\
w/o MaxAttentive-Matching  & 87.98 \\
\midrule
\midrule
Full Model & \textbf{88.69} \\
\bottomrule
\end{tabular}
\caption{Ablation studies on the dev set.}
\label{tab:ablation}
\end{table}

\subsection{Experiments on Paraphrase Identification}
\label{subsec:parahprase}
In this Sub-section, we compare our model with state-of-the-art models on the paraphrase identification task. We still experiment on the ``Quora Question Pairs'' dataset, and use the same dataset partition as Sub-section \ref{subsec:property}. 
This dataset is a brand-new dataset, and no previous results have been published yet. 
Therefore, we implemented three types of baseline models. 

\textbf{First}, under the Siamese framework, we implement two baseline models: ``Siamese-CNN'' and ``Siamese-LSTM''. 
Both of the two models encode two input sentences into sentence vectors with a neural network encoder, and make a decision based on the cosine similarity between the two sentence vectors. 
But they implement the sentence encoder with a CNN and a LSTM respectively.
We design the CNN and the LSTM model according to the architectures in \cite{wang2016semi}.

\textbf{Second}, based on the two baseline models, we implement two more baseline models ``Multi-Perspective-CNN'' and ``Multi-Perspective-LSTM''. In these two models, we change the cosine similarity calculation layer with our multi-perspective cosine matching function in  Eq.(\ref{equ:MP_cosine}), and apply a fully-connected layer (with $sigmoid$ function on the top) to make the prediction.

\textbf{Third}, we re-implement the ``L.D.C.'' model proposed by \cite{wang2016sentence}, which is a model under the ``matching-aggregation'' framework and acquires the state-of-the-art performance on several tasks.

\begin{table}[tbp]
\centering
\begin{tabular}{lc}
\toprule
Models                   & Accuracy\\
\midrule
Siamese-CNN & 79.60 \\
Multi-Perspective-CNN & 81.38 \\
Siamese-LSTM & 82.58 \\
Multi-Perspective-LSTM & 83.21 \\
L.D.C.  & 85.55 \\
\midrule
BiMPM & \textbf{88.17} \\
\bottomrule
\end{tabular}
\caption{Performance for paraphrase identification on the Quora dataset.}
\label{tab:quora}
\end{table}

Table \ref{tab:quora} shows the performances of all baseline models and our ``BiMPM'' model. We can see that ``Multi-Perspective-CNN'' (or ``Multi-Perspective-LSTM'') works much better than ``Siamese-CNN'' (or ``Siamese-LSTM''), which further indicates that our multi-perspective cosine matching function (Eq.(\ref{equ:MP_cosine})) is very effective for matching vectors. Our ``BiMPM'' model outperforms the ``L.D.C.'' model by more than two percent. Therefore, our model is very effective for the paraphrase identification task.

\subsection{Experiments on Natural Language Inference}
\label{subsec:inference}
In this Sub-section, we evaluate our model on the natural language inference task over the SNLI dataset \cite{bowman2015large}. 
We test four variations of our model on this dataset, where ``Only $P \rightarrow Q$'' and ``Only $P \leftarrow Q$'' are the single direction matching models described in Sub-section \ref{subsec:property}, ``BiMPM'' is our full model, and ``BiMPM (Ensemble)'' is an ensemble version of our ``BiMPM'' model. We design the ensemble model by simply averaging the probability distributions \cite{peng2015piefa,peng2017toward} of four ``BiMPM'' models, and each of the ``BiMPM'' model has the same architecture, but is initialized with a different seed.

\begin{table}[tbp]
\centering
\begin{tabular}{lc}
\toprule
Models                   & Accuracy \\
\midrule
\cite{bowman2015large} & 77.6\\
\cite{vendrov2015order} & 81.4 \\
\cite{mou2015natural} & 82.1 \\
\cite{rocktaschel2015reasoning} & 83.5 \\
\cite{liu2016learning} & 85.0 \\
\cite{liu2016modelling} & 85.1 \\
\cite{wang2015learning} & 86.1 \\
\cite{cheng2016long} & 86.3 \\
\cite{parikh2016decomposable} & 86.8 \\
\cite{munkhdalai2016neural} & 87.3 \\
\cite{sha2016OLING} & 87.5 \\
\cite{chen2016enhancing} (Single) & 87.7 \\
\cite{chen2016enhancing} (Ensemble) & 88.3 \\
\midrule
Only $P \rightarrow Q$ & 85.6 \\
Only $P \leftarrow Q$  & 86.3 \\
BiMPM  & 86.9 \\
BiMPM (Ensemble)  & \textbf{88.8} \\
\bottomrule
\end{tabular}
\caption{Performance for natural language inference on the SNLI dataset.}
\label{tab:snli}
\end{table}

Table \ref{tab:snli} shows the performances of the state-of-the-art models and our models. 
First, we can see that ``Only $P \leftarrow Q$'' works significantly better than ``Only $P \rightarrow Q$'', which tells us that, for natural language inference, matching the hypothesis against the premise is more effective than the other way around. 
Second, our ``BiMPM'' model works much better than ``Only $P \leftarrow Q$'', which reveals that matching premise against the hypothesis can also bring some benefits.
Finally, comparing our models with all the state-of-the-art models, we can observe that our single model ``BiMPM'' is on par with the state-of-the-art single models, and our `BiMPM (Ensemble)'' works much better than ``\cite{chen2016enhancing} (Ensemble)''. 
Therefore, our models achieve the state-of-the-art performance in both single and ensemble scenarios for the natural language inference task.

\subsection{Experiments on Answer Sentence Selection}
\label{subsec:answer-selection}
In this Sub-section, we study the effectiveness of our model for answer sentence selection tasks. The answer sentence selection task is to rank a list of candidate answer sentences based on their similarities to the question, and the performance is measured by the mean average precision (MAP) and mean reciprocal rank (MRR). We experiment on two datasets: TREC-QA \cite{wang2007jeopardy} and WikiQA \cite{yang2015wikiqa}. Experimental results of the state-of-the-art models \footnote{\cite{rao2016noise} pointed out that there are two versions of TREC-QA dataset: raw-version and clean-version. In this work, we utilized the clean-version. Therefore, we only compare with approaches reporting performance on this dataset.} and our ``BiMPM'' model are listed in Table \ref{tab:answer-selection}, where the performances are evaluated with the standard trec\_eval-8.0 script \footnote{http://trec.nist.gove/trec\_eval/}. We can see that the performance from our model is on par with the state-of-the-art models. Therefore, our model is also effective for answer sentence selection tasks.

\begin{table}[tbp]
\centering
\begin{tabular}{lcccc}
\toprule
\multirow{2}{*}{Models} & \multicolumn{2}{c}{TREC-QA} & \multicolumn{2}{c}{WikiQA} \\
                        & MAP          & MRR         & MAP          & MRR         \\
\midrule
\cite{yang2015wikiqa} & 0.695&0.763 & 0.652&0.665 \\
\cite{tan2015lstm} & 0.728  &	0.832 & -- & -- \\
Wang and Itty. \shortcite{wang2015faq} & 0.746  &	0.820 & -- & -- \\
\cite{santos2016attentive} & 0.753 	& 0.851 & 0.689 & 0.696 \\
\cite{yin2015abcnn} & --& --& 0.692 & 0.711 \\
\cite{miao2016neural} & --& --& 0.689 & 0.707 \\
\cite{wang2016sentence} & 0.771 & 0.845 & 0.706 & 0.723 \\
\cite{he2016pairwise} & 0.777&0.836 & 0.709&0.723 \\
\cite{rao2016noise} & 0.801& \textbf{0.877} & 0.701&0.718 \\
\cite{wang2016inner} & --&--& 0.734&0.742 \\
\cite{wang2016compare} & --&--& \textbf{0.743}&\textbf{0.755} \\
\midrule
BiMPM & \textbf{0.802} &0.875&0.718&0.731\\
\bottomrule
\end{tabular}
\caption{Performance for answer sentence selection on TREC-QA and WikiQA datasets.}
\label{tab:answer-selection}
\end{table}

\section{Related Work}
Natural language sentence matching (NLSM) has been studied for many years. Early approaches focused on designing hand-craft features to capture n-gram overlapping, word reordering and syntactic alignments phenomena \cite{heilman2010tree,wang2015faq}. This kind of method can work well on a specific task or dataset, but it's hard to generalize well to other tasks.
 
With the availability of large-scale annotated datasets~\cite{bowman2015large}, many deep learning models were proposed for NLSM. The first kind of framework is based the Siamese architecture \cite{bromley1993signature}, where sentences are encoded into sentence vectors based on some neural network encoders, and then the relationship between two sentences was decided solely based on the two sentence vectors \cite{bowman2015large,yang2015wikiqa,tan2015lstm}. However, this kind of framework ignores the fact that the lower level interactive features between two sentences are indispensable. Therefore, many neural network models were proposed to match sentences from multiple level of granularity \cite{yin2015abcnn,wang2016compare,wang2016sentence}. Experimental results on many tasks have proofed that the new framework works significantly better than the previous methods. Our model also belongs to this framework, and we have shown its effectiveness in Section \ref{sec:exper}.

\section{Conclusion}
In this work, we propose a bilateral multi-perspective matching (BiMPM) model under the ``matching-aggregation'' framework. 
Different from the previous ``matching-aggregation'' approaches, our model matches sentences $P$ and $Q$ in two directions ($P \rightarrow Q$ and $P \leftarrow Q$). And, in each individual direction, our model matches the two sentences from multiple perspectives.
We evaluated our model on three tasks: paraphrase identification, natural language inference and answer sentence selection.
Experimental results on standard benchmark datasets show that our model achieves the state-of-the-art performance on all tasks.


\bibliographystyle{named}
\bibliography{ijcai17}

\begin{thebibliography}{}

\bibitem[\protect\citeauthoryear{Bowman \bgroup \em et al.\egroup
  }{2015}]{bowman2015large}
Samuel~R Bowman, Gabor Angeli, Christopher Potts, and Christopher~D Manning.
\newblock A large annotated corpus for learning natural language inference.
\newblock {\em arXiv preprint arXiv:1508.05326}, 2015.

\bibitem[\protect\citeauthoryear{Bromley \bgroup \em et al.\egroup
  }{1993}]{bromley1993signature}
Jane Bromley, James~W. Bentz, L{\'e}on Bottou, Isabelle Guyon, Yann LeCun,
  Cliff Moore, Eduard S{\"a}ckinger, and Roopak Shah.
\newblock Signature verification using a "siamese" time delay neural network.
\newblock {\em IJPRAI}, 7(4):669--688, 1993.

\bibitem[\protect\citeauthoryear{Chen \bgroup \em et al.\egroup
  }{2016}]{chen2016enhancing}
Qian Chen, Xiaodan Zhu, Zhenhua Ling, Si~Wei, and Hui Jiang.
\newblock Enhancing and combining sequential and tree lstm for natural language
  inference.
\newblock {\em arXiv preprint arXiv:1609.06038}, 2016.

\bibitem[\protect\citeauthoryear{Cheng \bgroup \em et al.\egroup
  }{2016}]{cheng2016long}
Jianpeng Cheng, Li~Dong, and Mirella Lapata.
\newblock Long short-term memory-networks for machine reading.
\newblock {\em arXiv preprint arXiv:1601.06733}, 2016.

\bibitem[\protect\citeauthoryear{He and Lin}{2016}]{he2016pairwise}
Hua He and Jimmy Lin.
\newblock Pairwise word interaction modeling with deep neural networks for
  semantic similarity measurement.
\newblock In {\em NAACL}, 2016.

\bibitem[\protect\citeauthoryear{Heilman and Smith}{2010}]{heilman2010tree}
Michael Heilman and Noah~A Smith.
\newblock Tree edit models for recognizing textual entailments, paraphrases,
  and answers to questions.
\newblock In {\em NAACL}, 2010.

\bibitem[\protect\citeauthoryear{Hochreiter and
  Schmidhuber}{1997}]{hochreiter1997long}
Sepp Hochreiter and J{\"u}rgen Schmidhuber.
\newblock Long short-term memory.
\newblock {\em Neural computation}, 9(8):1735--1780, 1997.

\bibitem[\protect\citeauthoryear{Kingma and Ba}{2014}]{kingma2014adam}
Diederik Kingma and Jimmy Ba.
\newblock Adam: A method for stochastic optimization.
\newblock {\em arXiv preprint arXiv:1412.6980}, 2014.

\bibitem[\protect\citeauthoryear{LeCun \bgroup \em et al.\egroup
  }{2015}]{lecun2015deep}
Yann LeCun, Yoshua Bengio, and Geoffrey Hinton.
\newblock Deep learning.
\newblock {\em Nature}, 521(7553):436--444, 2015.

\bibitem[\protect\citeauthoryear{Liu \bgroup \em et al.\egroup
  }{2016a}]{liu2016modelling}
Pengfei Liu, Xipeng Qiu, and Xuanjing Huang.
\newblock Modelling interaction of sentence pair with coupled-lstms.
\newblock {\em arXiv preprint arXiv:1605.05573}, 2016.

\bibitem[\protect\citeauthoryear{Liu \bgroup \em et al.\egroup
  }{2016b}]{liu2016learning}
Yang Liu, Chengjie Sun, Lei Lin, and Xiaolong Wang.
\newblock Learning natural language inference using bidirectional lstm model
  and inner-attention.
\newblock {\em arXiv preprint arXiv:1605.09090}, 2016.

\bibitem[\protect\citeauthoryear{Miao \bgroup \em et al.\egroup
  }{2016}]{miao2016neural}
Yishu Miao, Lei Yu, and Phil Blunsom.
\newblock Neural variational inference for text processing.
\newblock In {\em ICML}, 2016.

\bibitem[\protect\citeauthoryear{Mikolov \bgroup \em et al.\egroup
  }{2013}]{mikolov2013distributed}
Tomas Mikolov, Ilya Sutskever, Kai Chen, Greg~S Corrado, and Jeff Dean.
\newblock Distributed representations of words and phrases and their
  compositionality.
\newblock In {\em Advances in neural information processing systems}, pages
  3111--3119, 2013.

\bibitem[\protect\citeauthoryear{Mou \bgroup \em et al.\egroup
  }{2015}]{mou2015natural}
Lili Mou, Rui Men, Ge~Li, Yan Xu, Lu~Zhang, Rui Yan, and Zhi Jin.
\newblock Natural language inference by tree-based convolution and heuristic
  matching.
\newblock {\em arXiv preprint arXiv:1512.08422}, 2015.

\bibitem[\protect\citeauthoryear{Munkhdalai and
  Yu}{2016}]{munkhdalai2016neural}
Tsendsuren Munkhdalai and Hong Yu.
\newblock Neural tree indexers for text understanding.
\newblock {\em arXiv preprint arXiv:1607.04492}, 2016.

\bibitem[\protect\citeauthoryear{Parikh \bgroup \em et al.\egroup
  }{2016}]{parikh2016decomposable}
Ankur~P Parikh, Oscar T{\"a}ckstr{\"o}m, Dipanjan Das, and Jakob Uszkoreit.
\newblock A decomposable attention model for natural language inference.
\newblock {\em arXiv preprint arXiv:1606.01933}, 2016.

\bibitem[\protect\citeauthoryear{Peng \bgroup \em et al.\egroup
  }{2015a}]{peng2015circle}
Xi~Peng, Junzhou Huang, Qiong Hu, Shaoting Zhang, Ahmed Elgammal, and Dimitris
  Metaxas.
\newblock From circle to 3-sphere: Head pose estimation by instance
  parameterization.
\newblock {\em Computer Vision and Image Understanding}, 136:92--102, 2015.

\bibitem[\protect\citeauthoryear{Peng \bgroup \em et al.\egroup
  }{2015b}]{peng2015piefa}
Xi~Peng, Shaoting Zhang, Yu~Yang, and Dimitris~N Metaxas.
\newblock Piefa: Personalized incremental and ensemble face alignment.
\newblock In {\em Proceedings of the IEEE International Conference on Computer
  Vision}, pages 3880--3888, 2015.

\bibitem[\protect\citeauthoryear{Peng \bgroup \em et al.\egroup
  }{2016}]{peng2016recurrent}
Xi~Peng, Rogerio~S Feris, Xiaoyu Wang, and Dimitris~N Metaxas.
\newblock A recurrent encoder-decoder network for sequential face alignment.
\newblock In {\em European Conference on Computer Vision}, pages 38--56.
  Springer International Publishing, 2016.

\bibitem[\protect\citeauthoryear{Peng \bgroup \em et al.\egroup
  }{2017}]{peng2017toward}
Xi~Peng, Shaoting Zhang, Yang Yu, and Dimitris~N Metaxas.
\newblock Toward personalized modeling: Incremental and ensemble alignment for
  sequential faces in the wild.
\newblock {\em International Journal of Computer Vision}, pages 1--14, 2017.

\bibitem[\protect\citeauthoryear{Pennington \bgroup \em et al.\egroup
  }{2014}]{pennington2014glove}
Jeffrey Pennington, Richard Socher, and Christopher~D Manning.
\newblock Glove: Global vectors for word representation.
\newblock In {\em EMNLP}, 2014.

\bibitem[\protect\citeauthoryear{Rao \bgroup \em et al.\egroup
  }{2016}]{rao2016noise}
Jinfeng Rao, Hua He, and Jimmy Lin.
\newblock Noise-contrastive estimation for answer selection with deep neural
  networks.
\newblock In {\em CIKM}, 2016.

\bibitem[\protect\citeauthoryear{Rockt{\"a}schel \bgroup \em et al.\egroup
  }{2015}]{rocktaschel2015reasoning}
Tim Rockt{\"a}schel, Edward Grefenstette, Karl~Moritz Hermann, Tom{\'a}{\v{s}}
  Ko{\v{c}}isk{\`y}, and Phil Blunsom.
\newblock Reasoning about entailment with neural attention.
\newblock {\em arXiv preprint arXiv:1509.06664}, 2015.

\bibitem[\protect\citeauthoryear{Santos \bgroup \em et al.\egroup
  }{2016}]{santos2016attentive}
Cicero~dos Santos, Ming Tan, Bing Xiang, and Bowen Zhou.
\newblock Attentive pooling networks.
\newblock {\em arXiv preprint arXiv:1602.03609}, 2016.

\bibitem[\protect\citeauthoryear{Sha \bgroup \em et al.\egroup
  }{2016}]{sha2016OLING}
Lei Sha, Baobao Chang, Zhifang Sui, and Sujian Li.
\newblock Reading and thinking: Re-read lstm unit for textual entailment
  recognition.
\newblock In {\em COLING}, 2016.

\bibitem[\protect\citeauthoryear{Tan \bgroup \em et al.\egroup
  }{2015}]{tan2015lstm}
Ming Tan, Cicero~dos Santos, Bing Xiang, and Bowen Zhou.
\newblock Lstm-based deep learning models for non-factoid answer selection.
\newblock {\em arXiv preprint arXiv:1511.04108}, 2015.

\bibitem[\protect\citeauthoryear{Vendrov \bgroup \em et al.\egroup
  }{2015}]{vendrov2015order}
Ivan Vendrov, Ryan Kiros, Sanja Fidler, and Raquel Urtasun.
\newblock Order-embeddings of images and language.
\newblock {\em arXiv preprint arXiv:1511.06361}, 2015.

\bibitem[\protect\citeauthoryear{Wang and Ittycheriah}{2015}]{wang2015faq}
Zhiguo Wang and Abraham Ittycheriah.
\newblock Faq-based question answering via word alignment.
\newblock {\em arXiv preprint arXiv:1507.02628}, 2015.

\bibitem[\protect\citeauthoryear{Wang and Jiang}{2015}]{wang2015learning}
Shuohang Wang and Jing Jiang.
\newblock Learning natural language inference with lstm.
\newblock {\em arXiv preprint arXiv:1512.08849}, 2015.

\bibitem[\protect\citeauthoryear{Wang and Jiang}{2016}]{wang2016compare}
Shuohang Wang and Jing Jiang.
\newblock A compare-aggregate model for matching text sequences.
\newblock {\em arXiv preprint arXiv:1611.01747}, 2016.

\bibitem[\protect\citeauthoryear{Wang \bgroup \em et al.\egroup
  }{2007}]{wang2007jeopardy}
Mengqiu Wang, Noah~A Smith, and Teruko Mitamura.
\newblock What is the jeopardy model? a quasi-synchronous grammar for qa.
\newblock In {\em EMNLP}, 2007.

\bibitem[\protect\citeauthoryear{Wang \bgroup \em et al.\egroup
  }{2016a}]{wang2016inner}
Bingning Wang, Kang Liu, and Jun Zhao.
\newblock Inner attention based recurrent neural networks for answer selection.
\newblock In {\em ACL}, 2016.

\bibitem[\protect\citeauthoryear{Wang \bgroup \em et al.\egroup
  }{2016b}]{wang2016multi}
Zhiguo Wang, Haitao Mi, Wael Hamza, and Radu Florian.
\newblock Multi-perspective context matching for machine comprehension.
\newblock {\em arXiv preprint arXiv:1612.04211}, 2016.

\bibitem[\protect\citeauthoryear{Wang \bgroup \em et al.\egroup
  }{2016c}]{wang2016semi}
Zhiguo Wang, Haitao Mi, and Abraham Ittycheriah.
\newblock Semi-supervised clustering for short text via deep representation
  learning.
\newblock In {\em CoNLL}, 2016.

\bibitem[\protect\citeauthoryear{Wang \bgroup \em et al.\egroup
  }{2016d}]{wang2016sentence}
Zhiguo Wang, Haitao Mi, and Abraham Ittycheriah.
\newblock Sentence similarity learning by lexical decomposition and
  composition.
\newblock In {\em COLING}, 2016.

\bibitem[\protect\citeauthoryear{Yang \bgroup \em et al.\egroup
  }{2015}]{yang2015wikiqa}
Yi~Yang, Wen-tau Yih, and Christopher Meek.
\newblock Wikiqa: A challenge dataset for open-domain question answering.
\newblock In {\em EMNLP}, 2015.

\bibitem[\protect\citeauthoryear{Yin \bgroup \em et al.\egroup
  }{2015}]{yin2015abcnn}
Wenpeng Yin, Hinrich Sch{\"u}tze, Bing Xiang, and Bowen Zhou.
\newblock Abcnn: Attention-based convolutional neural network for modeling
  sentence pairs.
\newblock {\em arXiv preprint arXiv:1512.05193}, 2015.

\end{thebibliography}

\end{document}